\documentclass[10pt,dvipsnames, twocolumn,letterpaper]{article}
\pdfoutput=1 
\usepackage[accsupp]{axessibility}
\usepackage{array}
\usepackage{graphicx}
\usepackage{amsmath}
\usepackage{amssymb}
\usepackage{booktabs}
\usepackage{multirow}
\usepackage{times}
\usepackage{epsfig}
\usepackage{float}
\usepackage{adjustbox}
\usepackage{array} 
\usepackage{enumitem}
\usepackage{caption}
\usepackage{makecell}
\usepackage{utfsym}
\usepackage{bbding}
\usepackage{xcolor}
\usepackage{pifont}
\usepackage{wrapfig}
\usepackage{listings}

\newcommand{\model}{PixelLM}

\usepackage[linesnumbered,ruled,vlined]{algorithm2e}
\usepackage{wrapfig}

\SetKwComment{Comment}{\color{green!50!black}\# }{}

\newcommand{\var}{\texttt}

\SetKwProg{Function}{def}{:}{}

\SetKwProg{For}{for}{:}{}
\SetKwProg{If}{if}{:}{}
\usepackage{cvpr}              
\definecolor{golden}{RGB}{171, 107, 35}

\definecolor{cvprblue}{rgb}{0.21,0.49,0.74}
\usepackage[pagebackref,breaklinks,colorlinks,citecolor=cvprblue,urlcolor=RoyalBlue,]{hyperref}

\def\confName{CVPR}
\def\confYear{2024}

\newcommand{\lisarec}{LISA$_{\text{rec}}$}
\newcommand{\lisaaug}{LISA$_{\text{aug}}$}
\newlength\savewidth\newcommand\shline{\noalign{\global\savewidth\arrayrulewidth
  \global\arrayrulewidth 1pt}\hline\noalign{\global\arrayrulewidth\savewidth}}

\usepackage{etoolbox}
\makeatletter
\AfterEndEnvironment{algorithm}{\let\@algcomment\relax}
\AtEndEnvironment{algorithm}{\kern2pt\hrule\relax\vskip3pt\@algcomment}
\let\@algcomment\relax
\newcommand\algcomment[1]{\def\@algcomment{\footnotesize#1}}
\renewcommand\fs@ruled{\def\@fs@cfont{\bfseries}\let\@fs@capt\floatc@ruled
  \def\@fs@pre{\hrule height.8pt depth0pt \kern2pt}%
  \def\@fs@post{}%
  \def\@fs@mid{\kern2pt\hrule\kern2pt}%
  \let\@fs@iftopcapt\iftrue}
\makeatother

\title{\includegraphics[width=0.04\linewidth]{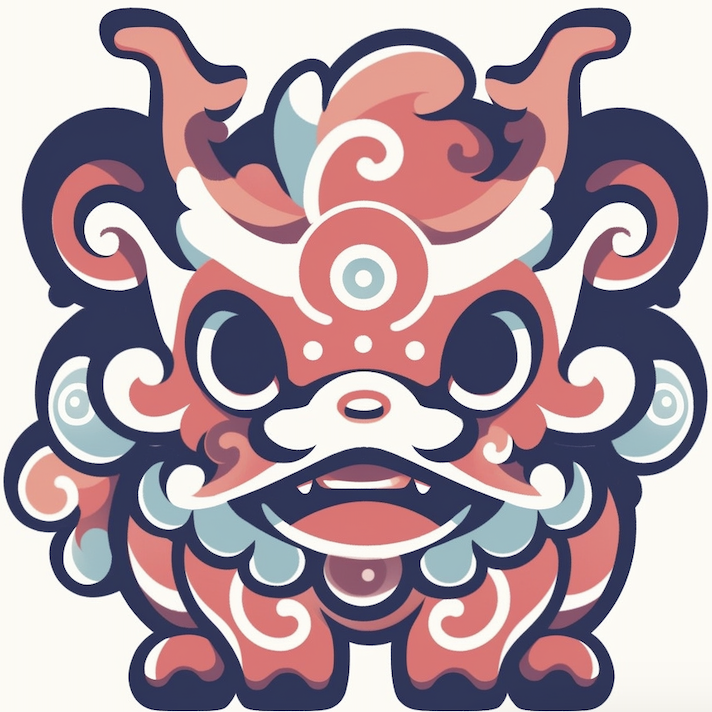} \model: Pixel Reasoning with Large Multimodal Model \vspace{-0.5em}}

\author{Zhongwei Ren$^{1}$\textsuperscript{\textnormal{*}},~~Zhicheng Huang$^{2}$\textsuperscript{\textnormal{*}},~~Yunchao Wei$^{1,4}$\textsuperscript{\textnormal{$\dagger$}},~~Yao Zhao$^{1,4}$,\\~~Dongmei Fu$^{2}$,~~Jiashi Feng$^{3}$,~~Xiaojie Jin$^{3}$\textsuperscript{*\textnormal{$\dagger$}\textnormal{$\ddagger$}}\\
{\fontsize{10.5pt}{12pt}\selectfont $^{1}$Beijing Jiaotong University $^{2}$University of Science and Technology Beijing} \\ {\fontsize{10.5pt}{12pt}\selectfont$^{3}$ByteDance Inc. $^{4}$Peng Cheng Laboratory}\\
{\hypersetup{urlcolor=golden}
  \fontsize{10pt}{12pt}\selectfont \href{https://pixellm.github.io/}{https://pixellm.github.io/}}
}

\begin{document}

\maketitle

\begin{abstract}
\vspace{-1em}
\footnotetext[1]{Equal contribution. Work done when Zhongwei and Zhicheng interned at ByteDance Inc.
   $^\dagger$Correspondence to Xiaojie Jin $<$\url{jinxiaojie@bytedance.com}$>$ and Yunchao Wei $<$\url{yunchao.wei@bjtu.edu.cn}$>$. \\$\ddagger$ Project lead.}
While large multimodal models (LMMs) have achieved remarkable progress, generating pixel-level masks for image reasoning tasks involving multiple open-world targets remains a challenge. To bridge this gap, we introduce \textbf{PixelLM}, an effective and efficient LMM for pixel-level reasoning and understanding. Central to PixelLM is a novel, lightweight pixel decoder and a comprehensive segmentation codebook. The decoder efficiently produces masks from the hidden embeddings of the codebook tokens, which encode detailed target-relevant information. With this design, PixelLM harmonizes with the structure of popular LMMs and avoids the need for additional costly segmentation models. Furthermore, we propose a target refinement loss to enhance the model's ability to differentiate between multiple targets, leading to substantially improved mask quality. To advance research in this area, we construct MUSE, a high-quality multi-target reasoning segmentation benchmark. PixelLM excels across various pixel-level image reasoning and understanding tasks, outperforming well-established methods in multiple benchmarks, including MUSE, single-
and multi-referring segmentation. Comprehensive ablations confirm the efficacy of each proposed component. All code, models, and datasets will be publicly available.

\vspace{-1.5em}
\end{abstract}    

\section{Introduction}
\label{sec:intro}
Built upon the success of Large Language Models (LLMs)~\cite{chatgpt,vicuna,openai2023gpt4,zhang2023llama}, large multimodal models (LMMs) have significantly enhanced high-level visual perception and user interaction experiences~\cite{li2023blip,zhu2023minigpt,liu2023llava,Qwen-VL}. Yet, most of them generate textual descriptions for global images or regions, with limited capability for pixel-level responses like object masks. This research gap limits the practical application of multimodal systems in fine-grained tasks, such as image editing, autonomous driving, and robotics.

Recent work~\cite{lai2023lisa} explores using LLMs to produce object masks in a novel reasoning segmentation task, which is more challenging and flexible for real-world applications. In contrast to traditional segmentation which explicitly specifies objects (e.g., ``orange''), reasoning segmentation requires complex reasoning for more intricate instructions (e.g., ``the fruit high in Vitamin-C''), which aligns well with the capabilities of LMMs. However, this method has two major drawbacks: 1) it cannot handle tasks involving multiple target objects, which are indispensable in real-world scenarios, and 2) it depends on a pre-trained image segmentation model like SAM~\cite{kirillov2023segment}. This reliance incurs substantial computational demands and confines the overall model's performance to the capability of segmentation model, consequently impeding the model's potential to enhance its performance through further training scaling up.

In this paper, we introduce \textbf{PixelLM}, an effective and efficient LMM for pixel-level reasoning and understanding. PixelLM proficiently handles tasks with an arbitrary number of open-set targets and diverse reasoning complexities. Its design maintains the fundamental structure of LMMs while avoiding additional, costly segmentation models, enhancing both efficiency and transferability to diverse applications. Fig.~\ref{fig:results_show} showcases PixelLM's capabilities in diverse segmentation tasks, producing high-quality target masks.

At the core of PixelLM is a novel pixel decoder and a segmentation codebook. The codebook contains learnable tokens that encode contexts and knowledge pertinent to targets referencing at different visual scales. The pixel decoder then produces target masks based on the hidden embeddings from the codebook tokens in conjunction with image features. Thanks to this design, PixelLM can generate high-quality masks without external segmentation models, significantly boosting its efficiency. Furthermore, we propose a target refinement loss to enhance the model's capability of differentiating between multiple targets, thus further improving the mask quality.

To facilitate model training and evaluation in this research area, we construct MUSE, a comprehensive multi-target reasoning segmentation dataset. Utilizing a GPT-4V~\cite{openai2023gpt4}-aided data curation pipeline, we create 246k question-answer pairs, covering 0.9 million instances. Our extensive ablation studies confirm the dataset's effectiveness in cultivating the model’s pixel reasoning capabilities.

PixelLM's effectiveness is clearly demonstrated across a variety of benchmarks. It achieves state-of-the-art performance on benchmarks including MUSE, single- and multi-referring segmentation, significantly outperforming baseline models. Notably, in comparison to models that rely on external segmentation models, such as ~\cite{lai2023lisa}, PixelLM reduces computational costs by up to 50\% while either maintaining or improving performance.

In summary, our contributions are:
\begin{itemize}
    \item We present PixelLM, a novel LMM for pixel-level reasoning and understanding. It handles tasks with diverse reasoning complexities, maintaining high efficiency.
    \item We construct MUSE, a high-quality multi-target reasoning segmentation dataset, facilitating model training and evaluation in future research.
    \item PixelLM achieves state-of-the-art results across a diverse range of benchmarks, clearly demonstrating its superior efficacy and efficiency compared to competing methods.
\end{itemize}
\section{Related Work}

\begin{figure*}[t]
\includegraphics[width=1\linewidth]{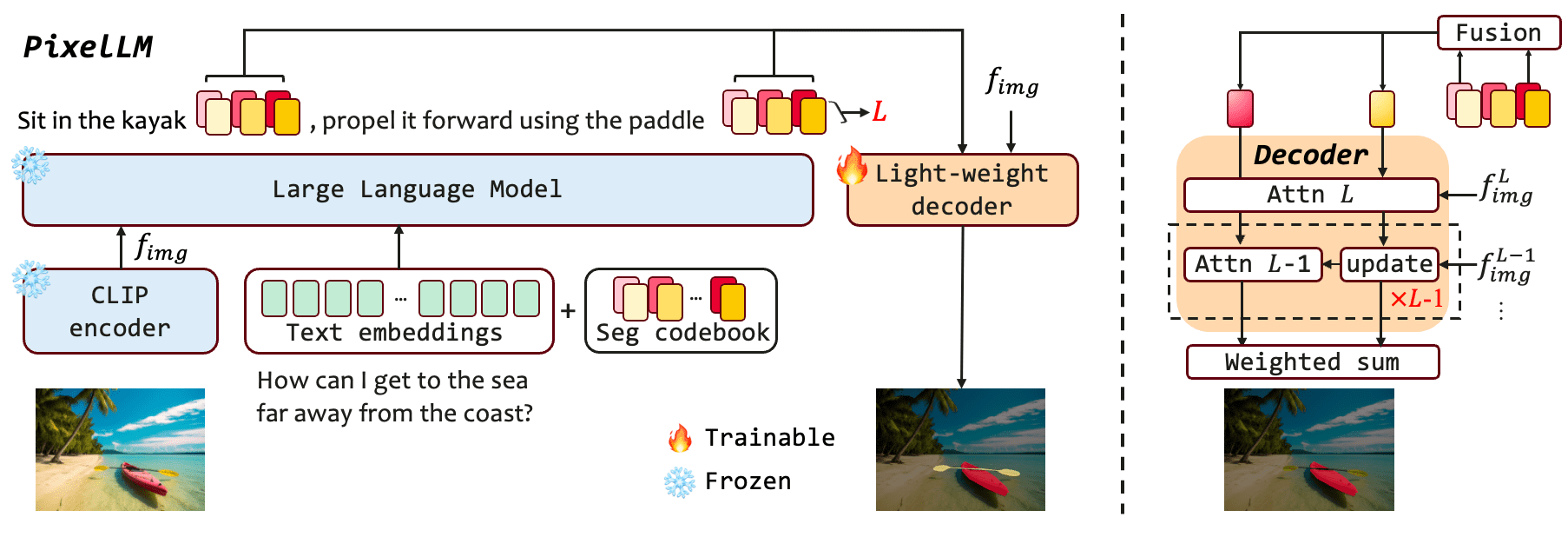}
\centering
\caption{Overview of the proposed PixelLM model architecture. (Left) Overall architecture. (Right) The proposed lightweight pixel decoder. Trainable LoRA parameters are incorporated into the LLM. All parameters except those for the CLIP encoder and LLM are trainable.}
\label{fig:overall}
\vspace{-1.5em}
\end{figure*}

\label{sec:related}
\subsection{Large Multimodal Models}
Large multimodal models (LMMs) have significantly enhanced the performance of tasks requiring the understanding of diverse modalities. These models generally fall into two categories based on their use of large language models (LLMs).

The first category~\cite{lu2022unified,wang2022ofa,yu2022coca} involves models trained from scratch or employing relatively smaller language models like BERT for text processing. They typically utilize a blend of contrastive and generative objectives to address a range of multimodal tasks. However, their limited language understanding capacity often hinders their performance in tasks that demand extensive world knowledge and reasoning abilities.

The advent of LLMs has spurred a new direction in LMM development, where LLMs are augmented with multimodal comprehension capabilities~\cite{zhang2022opt,touvron2023llama,alpaca, peng2023instruction}. Common approaches in this framework involve integrating adapters to align visual and textual representations within LLMs. Notable examples include BLIP-2~\cite{li2023blip}, Flamingo~\cite{alayrac2022flamingo}, MiniGPT-4~\cite{zhu2023minigpt}, llama-adapter~\cite{zhang2023llama,llama-adapter-v2}, LLaVA~\cite{liu2023llava}, InstructBLIP~\cite{instructblip}, InternGPT~\cite{2023interngpt}, and Qwen-VL~\cite{Qwen-VL}. While these methods have shown improved performance in vision-language tasks through instructional tuning, their primary limitation lies in generating only textual outputs about the entire image, thus constraining their utility in tasks requiring more granular, region-level or pixel-level understanding.

\subsection{Fine-Grained LMMs}
In many practical applications, understanding visual inputs at a more detailed level is crucial, such as specific regions or even at the pixel level. Several methods have been proposed to endow LLMs with this fine-grained understanding capability. GPT4RoI~\cite{zhang2023gpt4roi}, Shikra~\cite{chen2023shikra}, Vision-LLM~\cite{wang2023visionllm}, Kosmos-2~\cite{peng2023kosmos}, InternGPT~\cite{liu2023interngpt}, and Ferret~\cite{you2023ferret} offer grounding capabilities to specified image regions, typically encoding location coordinates as tokens for integration with LLMs. Unlike these methods, which lack the ability to generate pixel-wise masks, LISA~\cite{lai2023lisa} integrates SAM with LLMs for segmentation tasks. Moreover, LISA explores the use of LMMs for complex instruction reasoning, which differs from traditional tasks that rely on explicit human instructions for object or category identification. However, LISA is constrained to handling single targets in images, and the incorporation of SAM adds significant computational overhead. In contrast, our goal is to develop an efficient LMM capable of pixel-level image reasoning and understanding, accommodating a varied number of targets and diverse reasoning complexities.

\section{Method}
We first outline the framework in Sec.~\ref{subsec:arch}, elucidating two key designs: pixel decoder and segmentation codebook. Training objectives are introduced in Sec.~\ref{subsec:hardsample}.

\subsection{Model Design}
\label{subsec:arch}
\noindent \textbf{Framework overview}
As depicted in Fig.~\ref{fig:overall}, PixelLM features a streamlined architecture, comprising four main parts: \textit{i}) a pre-trained CLIP-ViT vision encoder $\mathcal{I}$ which aligns with text, \textit{ii}) a large language model $\mathcal{F}$, \textit{iii}) a lightweight pixel decoder $\mathcal{D}$ and  \textit{iv}) a segmentation codebook $C_{\text{seg}}$.  PixelLM processes image $x_{\text{img}}$ and query text $x_{\text{txt}}$, yielding interleaved text description and corresponding masks for varied number of targets.

While components \textit{i}) and \textit{ii}) adhere to well-established LMM architectures for ensuring compatibility, the pixel decoder and segmentation codebook are pivotal in empowering LMMs with mask generation capabilities across diverse scenarios. We utilize the segmentation codebook to encode target-relevant information into the embeddings of the codebook tokens, which the pixel decoder then transforms in conjunction with image features into precise masks. In the following, we detail their designs.

\noindent \textbf{Segmentation codebook} 
Aiming to enrich the encoding of target-specific information and thereby facilitate the generation of high-quality masks, we devise a comprehensive segmentation codebook. This codebook includes various groups of tokens, each representing different levels of granularity or scales in visual concepts, tailored to meet the demands of segmentation tasks. For example, proficient segmentation requires comprehension of both semantic categories and nuanced geometric shapes. These elements are typically represented at distinct network layers. In response, our single codebook integrates diverse visual information, effectively capturing both semantic and geometric aspects necessary for accurate segmentation. 

Specifically, the codebook consists of multiple token groups, each corresponding to a semantic scale of visual features from the image encoder. Formally, we define $C_{\text{seg}} = \left\{ c_n^{\ell} \in \mathbb{R}^d \right\}_{n=1,\ell=1}^{N,L}$, where $L$ and $N$ denote the number of visual scales and tokens per group, respectively, and $d$ represents the hidden dimension in LMMs. For clarity, we first set $N=1$ and introduce how the codebook tokens are integrated within the LMMs to encode requisite information for target mask generation.

For an input image $x_{\text{img}}$, the vision encoder $\mathcal{I}$ extracts a spectrum of multi-scale visual features $I_{\text{img}} = \{ I_{\text{img}}^{\ell} \}_{\ell=1}^{L}$ from $\mathcal{I}(x_{\text{img}})$, comprising $L$ visual features output at select layers of $\mathcal{I}$. The output of the final layer, $I_{\text{img}}^{L}$, encapsulates global image information and is transformed into the language space via a vision-to-language projection layer $p_{V\rightarrow T}$. Simultaneously, a vision-to-decoder projection $p_{V\rightarrow D}$ transforms all $I_{\text{img}}$ features, resulting in $f_{\text{img}} = \left\{f_{\text{img}}^{\ell}=p_{V\rightarrow D}(I_{\text{img}}^{\ell})\right\}_{\ell=1}^{L}$. The codebook tokens, combined with the input image and text, are then processed by the LLM to generate interleaved response $y_{\text{res}}$ in an auto-regressive way:
\begin{equation*}
y_{\text{res}}=\mathcal{F}(p_{V\rightarrow T}(I_{\text{img}}^{L}),x_{\text{txt}}, C_{\text{seg}}).
\end{equation*}
To help understand this process, consider an example of text query ``{\tt Segment the apple on the left}''. Then, the output $y_{\text{res}}$ contains $L$ tokens of $C_{\text{seg}}$: ``{\tt The apple is $c^1,\dots,c^L$}''. The corresponding hidden embeddings (i.e. the output of last layer of $\mathcal{F}$) of $C_{\text{seg}}$ are represented as $h=\left\{h^{\ell}\right\}_{\ell=1}^{L}$, which are inputs to the pixel decoder $\mathcal{D}$ alongside image features $f_{\text{img}}$ for mask generation. 

\begin{figure}[t]
\includegraphics[width=0.9\linewidth]{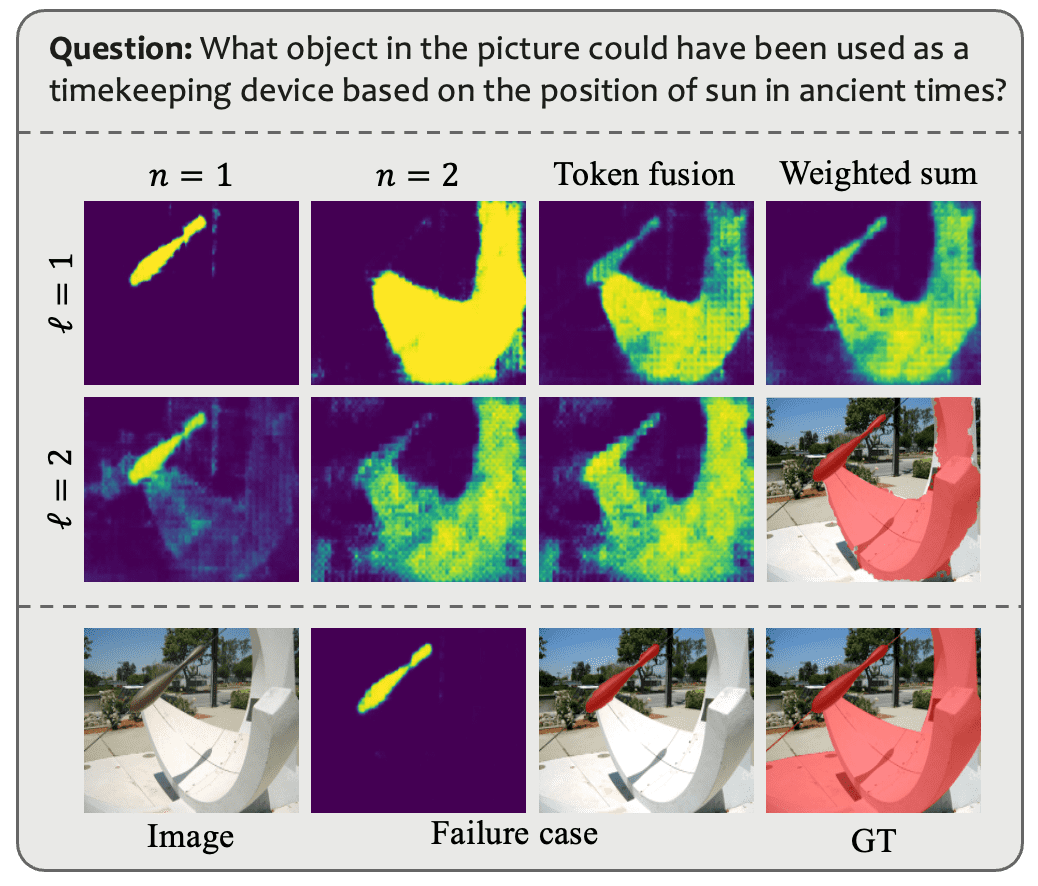}
\centering
\vspace{-1em}
\caption{The segmentation codebook example comprises two scales with two tokens each. Each attention map results from the interaction between one token and its corresponding image feature in the decoder. The first two rows depict the token fusion mechanism, while the final row demonstrates a failure case arising from the utilization of only one token.
}
\label{fig:multi_seg}
\vspace{-1em}
\end{figure}

We then explain the rationale behind setting $N>1$. As depicted in Fig.~\ref{fig:multi_seg}, scenarios involving multiple targets or increased complexity reveal that a single token may not adequately capture the full scope of target semantics, despite the LLM providing precise textual responses. To bolster the model's capacity for interpreting complex reasoning scenarios, we introduce multiple tokens within each scale group and perform a token fusion operation, denoted as $c^{\ell} = \left\{c_n^{\ell}\right\}_{n=1}^N$. Before the decoding process, a linear projection layer $\phi$ is utilized to convert the hidden states of these grouped tokens into a unified form, represented as $h^{\ell}=\phi(h_1^{\ell},\dots,h_N^{\ell})$. Fig.~\ref{fig:multi_seg} showcases the application of multiple tokens within each group. The post-decoding attention map visualizations demonstrate that individual tokens contribute distinct, yet complementary information, leading to a more effective mask generation than is possible with a single token approach.

\begin{figure*}[t]
\includegraphics[width=1\linewidth]{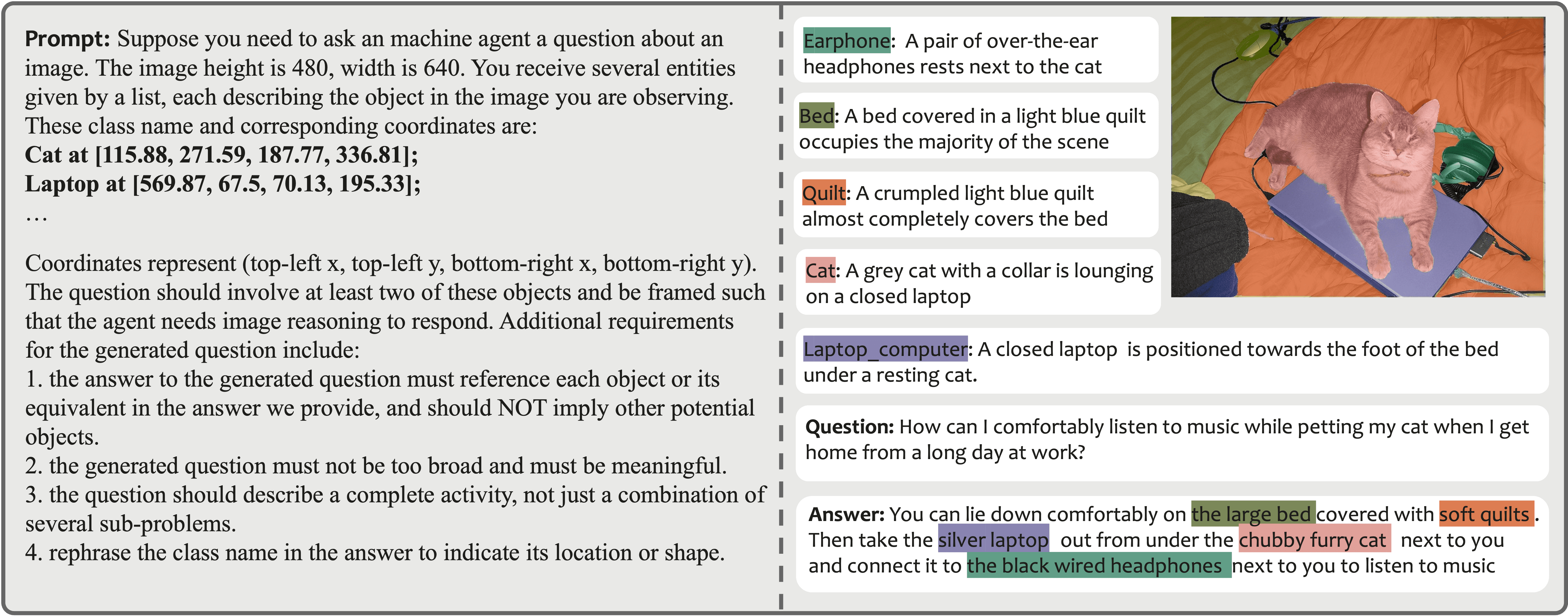}
\centering
\vspace{-1.5em}
\caption{The left panel illustrates the prompt employed in our GPT-4V data generation pipeline. The right panel showcases an example of the generated data.}
\label{fig:data_example}
\vspace{-1.5em}
\end{figure*}

\noindent \textbf{Pixel decoder} 
We design a novel and lightweight pixel decoder $\mathcal{D}$ to adeptly harness the multi-scale features from the vision encoder. This decoder is tasked with learning the transformation of these features, in conjunction with the hidden embeddings from $C_{\text{seg}}$, into precise segmentation masks. Such a design obviates the need for extra costly segmentation models like SAM~\cite{lai2023lisa}, thus significantly improving efficiency.

As depicted in the right panel of Fig.~\ref{fig:overall}, $\mathcal{D}$ consists of $L$ attention blocks $\left\{Attn^{\ell}\right\}_{\ell=1}^{L}$, each corresponding to distinct scales of image features and the codebook. For each targeted mask generation, $\mathcal{D}$ sequentially produces a mask score map $m^{\ell}$ at each scale $\ell$, 
which then directs the model’s attention to regions of higher relevance in the subsequent scale $\ell - 1$. This strategy works by guiding the model to focus on areas with high confidence scores in $m^{\ell}$, thereby facilitating more accurate mask generation.
\begin{equation}
\begin{aligned}
f_{img}^{\ell^{\prime}} &= \left\{ 
    \begin{array}{lc}
        f_{img}^L&\ell=L \\
        f_{img}^{\ell} \odot (\sigma(m^{\ell+1}) + 1) &\ell<L
    \end{array} \right. \\
m^{\ell}&=Attn^{\ell}(h^{\ell},f_{img}^{\ell^{\prime}}).
\end{aligned} 
\label{eq:decoder}
\end{equation}
\noindent where $f_{img}^{\ell^{\prime}}$ is the modulated feature at scale $\ell$, $\sigma$ is sigmoid function and $\odot$ is element-wise multiplication. Finally, we learn the weighting factors $\gamma=[\gamma^{\ell}]_{\ell=1}^{L}$ to combine mask maps at all scales to get the final segmentation result:
$\hat{M}=\sum_{\ell=1}^{L} \gamma^{\ell}m^{\ell}$ where $| \mathbf{\gamma} | = 1$. The detailed structure of the decoder is provided in Supp.~\ref{sec:supp_implement}.

\label{sec:method}




\subsection{Traning Objectives}

\label{subsec:hardsample}
\noindent \textbf{Target refinement loss} As the number of targets increases, the likelihood of the model encountering confusion and producing overlapping masks intensifies. To mitigate this issue, we introduce a target refinement loss. This simple yet effective strategy focuses on unclear pixels where multiple targets are predicted together. This helps the model in clearly identifying and learning different targets.

Denote the mask predictions as $\left\{\hat{M}_k \in \mathbb{R}^{H\times W}\right\}_{k=1}^{K}$ where $K$ is the total number of targets, $H$ and $W$ are the shape of mask. $\hat{M}_{k_i} \in \mathbb{R} $ represents the binary value of each pixel. Then we define a  map $A$ to assign increased weights to regions predicting multiple targets:
\begin{equation*}
    A_i = \left\{ 
    \begin{array}{lc}
        \alpha, \quad &\sum_k\hat{M}_{k_i}\geq2\\ 
        1, \quad &\sum_k\hat{M}_{k_i}<2
    \end{array}
\right.
\end{equation*}
where $\alpha$ is a hyper-parameter. The weighted loss is computed against the ground-truth $M_k$ for each mask as follows:
\begin{equation*}
    \mathcal{L}_{ref}=\frac{1}{KHW}\sum\nolimits_k\sum\nolimits_i A_i\mathcal{L}_{BCE}(\hat{M_{k_i}},M_{k_i})
\end{equation*}
where $\mathcal{L}_{BCE}$ is per-pixel binary cross-entropy loss.

\noindent\textbf{Overall loss}
The model is trained end-to-end using an auto-regressive cross-entropy loss $\mathcal{L}_{txt}$ for text generation (note the tokens from the segmentation codebook are also included in the loss calculation), a DICE loss $\mathcal{L}_{dice}$ and a target refinement loss $\mathcal{L}_{ref}$ for mask generation. The overall objective $\mathcal{L}$ constitutes the weighted sum of these losses, calibrated by $\lambda_{ref}$ and $\lambda_{dice}$:
\begin{equation*}
    \mathcal{L}=\mathcal{L}_{txt} + \lambda_{ref}\mathcal{L}_{ref} + \lambda_{dice}\mathcal{L}_{dice}
\end{equation*}

\section{Multi-target Reasoning Segmentation}
\label{sec:multi-reasoning}
Our objective is to develop a LLM capable of handling tasks involving an arbitrary number of open-set targets and diverse reasoning complexities. A primary challenge is the absence of a suitable dataset for model training.
We review existing public datasets and identify critical limitations: 1) Inadequate detail and object representation in segmentation masks; 2) A lack of question-answer pairs featuring complex reasoning and a varied number of objectives. 

To address these issues, we introduce a structured annotation pipeline for constructing the multi-target reasoning segmentation (MUSE) data. Examples of MUSE are shown in Fig.~\ref{fig:results_show}. MUSE stands out with its open-set concepts, detailed object descriptions, complex multi-target question-answer pairs, and instance-level mask annotations. In the next, we elaborate on the construction of MUSE.

\vspace{-1pt}
\subsection{MUSE Dataset}
\vspace{-1pt}
\label{subsec:datasetoverall}

Two public datasets, RefCOCO series~\cite{kazemzadeh2014referitgame} and ReasonSeg dataset~\cite{lai2023lisa}, are pertinent to our goal. RefCOCO guides segmentation with explicit target object names, e.g. ``orange'', lacking more complicated instructions, e.g. ``the fruit high in Vitamin-C''.
Besides, they also fall short in offering multi-target question-answer pairs with target descriptions directly connected to segmentation masks, which however is a common requirement in real-world scenarios, like ``How to make fruit salad?"

To this end, a total of 910k high-quality instance segmentation masks are selected from the LVIS dataset, along with detailed textual descriptions based on image content. Utilizing these instances, we construct 246k question-answer pairs, averaging 3.7 targets per answer. This dataset is then divided into three splits: {\tt train}, {\tt val}, and {\tt test}, containing 239k, 2.8k, and 4.3k question-answer pairs, respectively. The test split comprises two parts:  the number of targets involved in the question are less or more than three. Please see Supp.~\ref{subsec:data_ana} for more analysis of MUSE.




\vspace{-4pt}
\subsection{Dataset Generation Pipeline}
\label{subsec:pipeline}
\vspace{-3pt}

We first try to use GPT-4 to construct our dataset: using LLAVA~\cite{liu2023llava} for image captioning and then GPT-4 for generating questions about multiple image regions. We utilize images with pre-existing mask annotations to reduce annotation costs. The image caption, manually selected object names, and bounding box coordinates from the image are input into GPT-4 to facilitate answer selection and question formulation. However, due to the inability to directly perceive image content, the content of question-answer pairs generated by this method is often confined to the caption's description, significantly limiting the diversity of the data. 


To address these limitations, the pipeline is refined in two key ways. First, we convert to the more advanced GPT-4V which shows strong capabilities in understanding visual contents~\cite{openai2023gpt4}.
This model has been instrumental in generating more nuanced and naturalistic questions. Additionally, we implement a more dynamic approach for answer generation. Specifically, we feed all the instance category names and corresponding bounding box coordinates in the image to GPT-4V. Using carefully crafted prompts, GPT-4V autonomously selects instances to construct question-answer pairs relevant to the image content. An example of such a prompt is illustrated in Fig.~\ref{fig:data_example}. More details of data filtering and the data generation pipeline are provided in Supp.~\ref{sec:sup_muse}.

\subsection{Evaluation}
\label{subsec:evaluation}

To evaluate our task, we focus on three main aspects: \textit{i}) the generation of natural text descriptions aligned with corresponding object masks, \textit{ii}) the accuracy of the match between object masks and text descriptions, and \textit{iii}) the quality of the masks. Since we focus on the mask quality of each target, we do not evaluate image-level captioning.


The evaluation process for each question involves four steps: \textbf{First}, we match the predicted masks with ground-truth masks based on mask IoU scores using bipartite matching, similar to DETR~\cite{detr}. Any unassigned predictions or ground-truths are assigned an empty mask. \textbf{Second}, we replace the position of the mask in the generated text with the corresponding ground-truth object description. For example, as shown in Fig.~\ref{fig:overall}, the text might read: ``{\tt Sit in the kayak (a red kayak parked on the beach), propel it forward using the paddle (a double-bladed paddle on the kayak)}''. The content in brackets is a ground-truth description. \textbf{Third}, this modified text prompts GPT-3.5 to score each prediction from 1 to 10, with higher scores indicating better quality, and unassigned predictions receiving a score of 0. \textbf{Fourth}, the final score for each prediction is the product of the GPT and IoU scores, from which we calculate the gIoU and cIoU metrics. For more evaluation details, please refer to Supp.~\ref{sec:supp_implement}.

\vspace{-3pt}
\section{Experiment}
\label{sec:experiment}
\vspace{-2.5em}

\begin{table*}[t]\centering
    \footnotesize
    \scalebox{0.95}
    {
        \begin{tabular}{l|c|c|cc|cc|cc|cc }
            \toprule
            
            \multirow{3}*{Method} &\multirow{3}*{\makecell{w/o\\SAM}} &\multirow{3}*{TFLOPs} & \multicolumn{2}{c|}{Val} & \multicolumn{6}{c}{Test}   \\ 
            

            \specialrule{0em}{0pt}{1pt}
            \cline{4-11}
            \specialrule{0em}{0pt}{1pt}
            ~  & & & \multicolumn{2}{c|}{overall}  & \multicolumn{2}{c|}{few targets} & \multicolumn{2}{c|}{many targets} & \multicolumn{2}{c}{overall}  \\ 

            \specialrule{0em}{0pt}{1pt}
            \cline{4-11}
            \specialrule{0em}{0pt}{1pt}
            
            ~  & & & gIoU & cIoU & gIoU & cIoU &gIoU &cIoU &gIoU &cIoU    \\ 
            
            \specialrule{0em}{0pt}{1pt}
            \hline
            \specialrule{0em}{0pt}{1pt}

            

            {\color{lightgray}SEEM}~\citep{zou2023segment}& {\color{lightgray}\Checkmark} &{\color{lightgray}0.43}   &{\color{lightgray}13.6}  &{\color{lightgray}16.2} &{\color{lightgray}23.6} &{\color{lightgray}24.9} &{\color{lightgray}8.5} &{\color{lightgray}13.2}  &{\color{lightgray}11.7} &{\color{lightgray}15.7} \\
            
            \specialrule{0em}{0pt}{1pt}
            \hline
            \specialrule{0em}{0pt}{1pt}

            LISA-7B~\cite{lai2023lisa} &  \XSolidBrush &7.16 & 18.8 &29.0  &24.7 &36.5 &9.6 &24.5  &12.8  &27.1   \\

            LISA-7B$_{\text{rec}}$ &  \XSolidBrush &7.16   &24.5  &31.1 &30.0 &30.9 &12.4  &23.2  &16.2 &24.8  \\
            
             LISA-7B$_{\text{aug}}$ &  \XSolidBrush &7.16 &42.0 &46.1  &43.5  &52.0 &37.7 &42.3  &38.9 &44.4 \\


            PixelLM-7B$^{\dagger}$ &  \Checkmark &\textbf{3.57} &39.9  &48.0  &43.1  &56.7  &36.0  &38.2  &37.5 &42.2\\

            PixelLM-7B &  \Checkmark &\textbf{3.57} &\textbf{42.6}  &\textbf{50.7}  &\textbf{44.6} &\textbf{59.2}  &\textbf{37.7} &\textbf{42.8}  &\textbf{39.2} &\textbf{46.3}\\

            \specialrule{0em}{0pt}{1pt}
            \cline{1-11}
            \specialrule{0em}{0pt}{1pt}
            
            LISA-Llama2-13B~\cite{lai2023lisa} &  \XSolidBrush &10.24 &20.4  &29.2  &27.5  &38.5  &10.9  &25.6 &14.4 &28.4 \\
            LISA-Llama2-13B$_{\text{aug}}$ &  \XSolidBrush &10.24 &43.6  &50.2  &44.7  &60.0  &41.2  &\textbf{47.9}  &41.9 &50.5 \\
            
            

            PixelLM-Llama2-13B$^{\dagger}$ &  \Checkmark &\textbf{6.65} &43.0  &51.7  &44.8  &61.6  &39.3  &44.6 &40.5 &48.2  \\
            PixelLM-Llama2-13B &  \Checkmark &\textbf{6.65} &\textbf{44.8}  &\textbf{54.1}  &\textbf{45.2}  &\textbf{62.9}  &\textbf{41.5}  &47.6  &\textbf{42.3} &\textbf{51.0} \\
            \bottomrule            
        \end{tabular}
    }
    \vspace{-1em}
    \caption{\textbf{Comparison on MUSE benchmark.}  $^\dagger$ denotes PixelLM w/o using the token fusion mechanism and target refinement loss. 
      }
    \label{table:reason_seg}   
\vspace{-1.5em}
\end{table*}

\begin{center}

\begin{table}[t]
    \footnotesize

       \centering
\hspace{-3mm}
    \resizebox{\columnwidth}{!}
    {
 
        \begin{tabular}{l| p{0.3cm} | p{0.3cm} p{0.4cm} p{0.45cm} | p{0.3cm} p{0.4cm} p{0.45cm}  | p{0.4cm} p{0.6cm}  }
            \toprule
            
            \multirow{2}*{\hspace{-1.8mm}Method} & \multirow{2}*{\hspace{-1.5mm}\makecell{w/o\\SAM}} & \multicolumn{3}{c|}{MrefCOCO} & \multicolumn{3}{c|}{MrefCOCO+}  & \multicolumn{2}{c}{MrefCOCOg}  \\ 
            
            \specialrule{0em}{0pt}{1pt}
            \cline{3-10}
            \specialrule{0em}{0pt}{1pt}
            
            ~  & & val & testA & testB & val & testA & testB & val(U) & test(U)  \\

            \specialrule{0em}{0pt}{1pt}
            \hline
            \specialrule{0em}{0pt}{1pt}
             
            \hspace{-1.8mm}LISA~\cite{lai2023lisa} & \XSolidBrush &36.7&38.3&36.4&34.0&36.3&32.1&34.5&36.2 
            \\
            
            \hspace{-1.8mm}LISA$_{\text{aug}}$ &  \XSolidBrush & 68.9&   70.8    &66.3  &59.8  &62.2   &54.1  &62.3 &63.9  \\


            \hspace{-1.8mm}PixelLM$^{\dagger}$ &  \Checkmark & 70.3   &74.2    &66.2 & 64.4 & 69.6  &57.0 & 64.0     & 67.0  \\

            \hspace{-1.8mm}PixelLM &  \Checkmark & \textbf{72.7}  &  \textbf{76.2}    &\textbf{68.1}  & \textbf{65.7}  &\textbf{71.3}   &\textbf{57.7} & \textbf{65.8}  &   \textbf{67.7}   \\
            
            \bottomrule            
        \end{tabular}
        }
    
    \vspace{-1em}
    \caption{\textbf{Results on the multi-referring segmentation benchmark.} The meaning of $^{\dagger}$ is the same as Tab.~\ref{table:reason_seg}.}  
    \label{table:phrase_seg}   
\vspace{-1em}
\end{table}
\end{center}

\begin{table}[t]
    \footnotesize
    \centering
    \resizebox{\columnwidth}{!}
    {
        \begin{tabular}{p{1.6cm}|p{0.3cm} | p{0.3cm} p{0.4cm} p{0.45cm} | p{0.3cm} p{0.4cm} p{0.45cm}  | p{0.4cm} p{0.6cm}  }
            \toprule
            
            \multirow{2}*{\hspace{-1.8mm}Method} & \multirow{2}*{\hspace{-1.5mm}\makecell{w/o\\SAM}} & \multicolumn{3}{c|}{refCOCO} & \multicolumn{3}{c|}{refCOCO+}  & \multicolumn{2}{c}{refCOCOg}  \\ 
            
            \specialrule{0em}{0pt}{1pt}
            \cline{3-10}
            \specialrule{0em}{0pt}{1pt}
            
            ~  & & val & testA & testB & val & testA & testB & val(U) & test(U)  \\ 
            
            
            
            \specialrule{0em}{0pt}{1pt}
            \hline
            \specialrule{0em}{0pt}{1pt}
     
            \hspace{-1.8mm}MCN~\citep{luo2020multi} & \Checkmark& 62.4 & 64.2 & 59.7 & 50.6 & 55.0 & 44.7 & 49.2 & 49.4 \\

            \hspace{-1.8mm}VLT~\citep{ding2021vision} & \Checkmark& 67.5 & 70.5 & 65.2 & 56.3 & 61.0 & 50.1 & 55.0 & 57.7 \\

            \hspace{-1.8mm}CRIS~\citep{wang2022cris} & \Checkmark& 70.5 & 73.2 & 66.1 & 62.3 & 68.1 & 53.7 & 59.9 & 60.4 \\

            \hspace{-1.8mm}LAVT~\citep{yang2022lavt} & \Checkmark& 72.7 & 75.8 & 68.8 & 62.1 & 68.4 & 55.1 & 61.2 & 62.1 \\
            
            \hspace{-1.8mm}ReLA~\citep{liu2023gres} & \Checkmark& 73.8 & 76.5 & 70.2 & 66.0 & 71.0 & 57.7 & 65.0 & 66.0 \\
            
            \hspace{-1.8mm}X-Decoder~\citep{zou2023generalized} & \Checkmark& - & - & - & - & - & - & 64.6 & -  \\

            \hspace{-1.8mm}SEEM~\citep{zou2023segment} & \Checkmark& - & - & - & - & - & - & 65.7 & -    \\
            
            \specialrule{0em}{0pt}{1pt}
            \hline
            \specialrule{0em}{0pt}{1pt}
            
            \hspace{-1.8mm}LISA~\cite{lai2023lisa}  & \XSolidBrush & \textbf{74.1} & \textbf{76.5} & \textbf{71.1} & 62.4 & 67.4 & 56.5 & 66.4 & 68.5 
            \\
            
            \hspace{-1.8mm}LISA$_{\text{aug}}$ &  \XSolidBrush &74.0  &76.3  &70.4 &62.5  & 66.3&   56.0 &67.0   & 69.1  \\



            \hspace{-1.8mm}PixelLM  &  \Checkmark &73.0   &\textbf{76.5}    &68.2  &\textbf{66.3} & \textbf{71.7}    &\textbf{58.3} &\textbf{69.3}   & \textbf{70.5}   \\
            
            \bottomrule            
        \end{tabular}
    }
    \vspace{-1em}
    \caption{\textbf{Results on the referring segmentation benchmark.} }   
    \label{table:refer_seg}   
\vspace{-1.5em}
\end{table}

\begin{table*}
\footnotesize
    \centering
    \begin{subtable}[h]{0.5\textwidth}
        \setlength{\tabcolsep}{8pt}
        \centering
        \begin{tabular}{l | cc | cc}
    \multirow{2}*{Architecture} &\multicolumn{2}{c|}{MUSE Val} &\multicolumn{2}{c}{refCOCOg} \\
     & gIoU & cIoU &  val(U) & test(U) \\
    \shline
    baseline &40.1  &47.2 &64.3 &65.6\\
    
    + 2 scale feature-fusion & \underline{42.6} & 50.7 &69.3 &\underline{70.5} \\
    + 3 scale feature-fusion  &42.3  &\underline{51.4}  &\underline{69.8} &70.4 \\
    + w/ CLIP-H & \textbf{47.8} & \textbf{55.2} &\textbf{72.6} &\textbf{73.0} \\
\end{tabular} 
       \caption{Different number of scales and larger vision encoder.}
       \label{tab:abla_arch}
    \end{subtable}
     \hfill
    \begin{subtable}[h]{0.45\textwidth}
        \setlength{\tabcolsep}{8pt}
       
        \begin{tabular}{c | cc | cc}
    \multirow{2}*{\makecell{Token number $N$\\per-group}} &\multicolumn{2}{c|}{MUSE Val} &\multicolumn{2}{c}{refCOCOg} \\
     & gIoU & cIoU &  val(U) & test(U) \\
    \shline

    1 &40.7 &48.6 &68.0  &68.3  \\

    2 &42.0  &49.5 &69.0 &69.9 \\
    
    3 &42.6  &\textbf{50.7} &\textbf{69.3} &70.5  \\
    4 &\textbf{43.0}  &50.3  &69.1  &\textbf{70.6}\\
    
\end{tabular}
        \caption{ Different token number in each group.}
        \label{tab:abla_segnum}
        
     \end{subtable}

        \bigskip 
     \begin{subtable}[h]{0.5\textwidth}
        \centering
        \setlength{\tabcolsep}{9.5pt}

        \begin{tabular}{ c c | cc | cc}
                  \multirow{2}*{\makecell{Token \\fusion}} & \multirow{2}*{\makecell{Target\\refinement loss}}   & \multicolumn{2}{c|}{MUSE Val} &\multicolumn{2}{c}{refCOCOg} \\
                 
                  &    &gIoU  &cIoU &val(U)  &test(U)  \\
                \shline
                  &   &39.9 &48.0  &68.0  &68.3   \\
                 \Checkmark &  &41.5 &50.1 &69.3  &70.5    \\
                  & \Checkmark &40.7 &48.6 &68.0    &68.3   \\
                 \Checkmark & \Checkmark  &\textbf{42.6} &\textbf{50.7} & \textbf{69.3} & \textbf{70.5}  \\
            \end{tabular}
            \caption{Token fusion mechanism and target refinement loss.}
            \label{tab:abla_method}
    
     \end{subtable}  
     \hfill
     \begin{subtable}[h]{0.45\textwidth}
        \centering
        \setlength{\tabcolsep}{8pt}

        \begin{tabular}{c | c|cc }
    \multirow{2}*{\makecell{Data \\ generator}} &\multirow{2}*{\makecell{Data \\amount}} &\multicolumn{2}{c}{MUSE Val}  \\
    &  & gIoU & cIoU  \\
    \shline

    GPT-4 &30k &30.0 &35.7  \\
            \hline
    \multirow{3}*{GPT-4V} &30k &35.6 &40.0 \\
    
     &100k &38.9 &45.7  \\
     &200k &\textbf{42.6} &\textbf{50.7}  \\
    
\end{tabular}
        \caption{Different data generators and number of question-answer pairs.}
        \label{tab:abla_gpt}

    \end{subtable}
     \vspace{-2ex}
    \caption{\textbf{Ablations.} We conduct all experiments based on our 7B model. }
    \label{tab:abl_sum}
\vspace{-2em}
\end{table*}

In this section, we first present the implementation details and then show the comparison results on benchmarks. Finally, we ablate on the key components in PixelLM.

\subsection{Implementation Details}
\label{subsec:implement_detail}

We use pre-trained multimodal model from LlaVA-7B and LlaVA-llama2-13B, with LoRA adopted for efficient fine-tuning. The vision encoder uses a fixed CLIP-ViT-L/14-336 model, modified with linearly interpolated position embeddings for processing 448x resolution images. The trainable parts of our model include the pixel decoder $\mathcal{D}$, LoRA parameters, segmentation codebook $C_{\text{seg}}$, the vision-to-language and vision-to-decoder projection layers $p_{V\rightarrow T}$ and $p_{V\rightarrow D}$. To facilitate task evaluation, we convert refCOCO series datasets into multi-referring segmentation datasets (detailed in Supp.~\ref{subsec:multi_referring}). Training involves random sampling from multi-referring segmentation datasets (ADE20K~\cite{ade20k}, COCO-Stuff~\cite{caesar2018cvpr}, LVIS-PACO~\cite{ramanathan2023paco}, refCOCO series~\cite{kazemzadeh2014referitgame}, $\sim$200k images in total), VQA data (LLAVA-150k), and MUSE. The training steps follow LISA~\cite{lai2023lisa}, requiring approximately 1.5/2 days on 8 A100 GPUs for the 7B/13B model.

\subsection{Benchmarks and Baselines}
\noindent \textbf{Benchmarks} We evaluate PixelLM on three benchmarks with a varied number of targets: MUSE, multi-referring segmentation and the conventional referring segmentation (refCOCO series). The former two involve multiple targets and the last one focuses on a single object. Through this evaluation, we validate the versatility of PixelLM in diverse mask generation tasks. For the MUSE benchmark, we use the GPT-based evaluation metric as detailed in Sec.~\ref{subsec:evaluation}. In multi-referring segmentation, we formulate multi-target queries from the refCOCO dataset's annotations, asking models to segment 3 to 6 objects per image. These queries are structured as ``{\tt Please segment the $<$objects$>$}'', with $<$objects$>$ being comma-separated object descriptors. The responses involve generating masks corresponding to the listed objects' sequence.  For refCOCO series, we follow previous methods to calculate the gIoU and cIoU scores.

\noindent \textbf{Baselines} To our best knowledge, PixelLM is the first to handle complex pixel reasoning tasks involving multiple targets. To demonstrate the effectiveness of PixelLM, we establish four strong baselines for comparative analysis on the aforementioned benchmarks. Three baselines evolve from LISA~\cite{lai2023lisa}, which is the most relevant work with PixelLM, and one additional non-LLM based model.
\begin{itemize}
\item LISA~\cite{lai2023lisa}: The original model, designed for single-target segmentation, employs SAM for mask generation.
\item \lisarec: To address the limits of LISA, this variant employs a two-step approach: first identify target objects in text form and then ask LISA to segment them one by one. Specifically, LLAVA-13B extracts noun phrase responses from multi-target queries using the prompt: ``{\tt $<$question$>$, Please only use noun phrases in answer}'', which are then input to LISA for segmentation.
\item \lisaaug: An augmented version of LISA, adapted to generate masks for multiple targets and additionally trained with the MUSE dataset.
\item SEEM~\cite{zou2023segment}: A state-of-the-art image segmentation model, SEEM is limited to producing segmentation masks only. Hence, its evaluation is confined to mask quality assessment across each benchmark.
\end{itemize}

\subsection{Results on MUSE}
\vspace{-3pt}
Tab.~\ref{table:reason_seg} compares PixelLM with competing methods on the multi-target reasoning segmentation task. For SEEM and the original LISA, since they can only generate a single segmentation mask given query text, we report the IoU between prediction and the ground truth which merges all target masks. Although the text description for the target is ignored, thus leading to a simpler setting when compared with ours, we can observe their performance is still inferior, which demonstrates the challenges of this task. 
Although \lisarec~improves upon LISA by simplifying the generation task by first generating the target for multi-target queries, it still largely lags behind our models.
This highlights the importance of end-to-end training on the task. By training on MUSE, the performance of the augmented LISA \lisaaug~further improves on the dataset, which validates the benefits of our dataset in training an LMM with strong pixel reasoning capabilities. 


PixelLM stands out in both efficiency and performance. Under the fair setting involving equivalent training data and LMM sizes, PixelLM-7B and PixelLM-Llama2-13B consistently outperform their counterparts (\lisaaug~and LISA-Llama2-13B$_{\text{aug}}$, respectively) across almost all evaluated metrics. Notably, PixelLM eschews additional segmentation modules, thereby conferring substantial efficiency advantages. In terms of computational efficiency, PixelLM achieves a remarkable reduction in TFlops by 50\% and 35\% for the 7B and 13B model sizes, respectively, compared to LISA. Interestingly, while PixelLM benefits from model scaling up: PixelLM-Llama2-13B boosts upon the PixelLM with a 7B LMM, it still maintains smaller TFlops (6.65 vs. 7.16) than \lisaaug~ which includes an extra SAM model. Furthermore, we conduct an ablation study to ascertain the impact of the proposed token fusion mechanism and target refinement loss. PixelLM, devoid of these two components, is denoted as PixelLM$^\dagger$. The results clearly demonstrate that the complete PixelLM configuration significantly outperforms PixelLM$^\dagger$, thereby affirming the contribution of these components to the model's efficacy. This trend is consistent in the 13B model variant as well. Detailed ablation analyses of each component are presented in Sec.~\ref{subsec:ablation}.

\vspace{-3pt}
\subsection{Results on Referring Segmentation}
\vspace{-3pt}
Tab.~\ref{table:phrase_seg} presents the results on the multi-target referring segmentation dataset, wherein PixelLM demonstrates superior performance across all data splits, significantly outperforming LISA and its enhanced variant, \lisaaug. Notably, PixelLM exhibits enhanced efficiency compared to the LISA models, as evidenced by its lower TFlops (as detailed in Tab.~\ref{table:reason_seg}). The superiority of the token fusion and target refinement loss components within PixelLM is further corroborated through the comparison with PixelLM$^\dagger$.

We also compare results on the conventional single-target refCOCO series dataset. Despite the dataset's focus on single target segmentation, PixelLM attains commendable performance. As shown in Tab.~\ref{table:refer_seg}, PixelLM achieves the highest scores on most metrics, particularly excelling in the more challenging refCOCO+/refCOCOg datasets. 

\begin{figure}[t]
\includegraphics[width=0.9\linewidth]{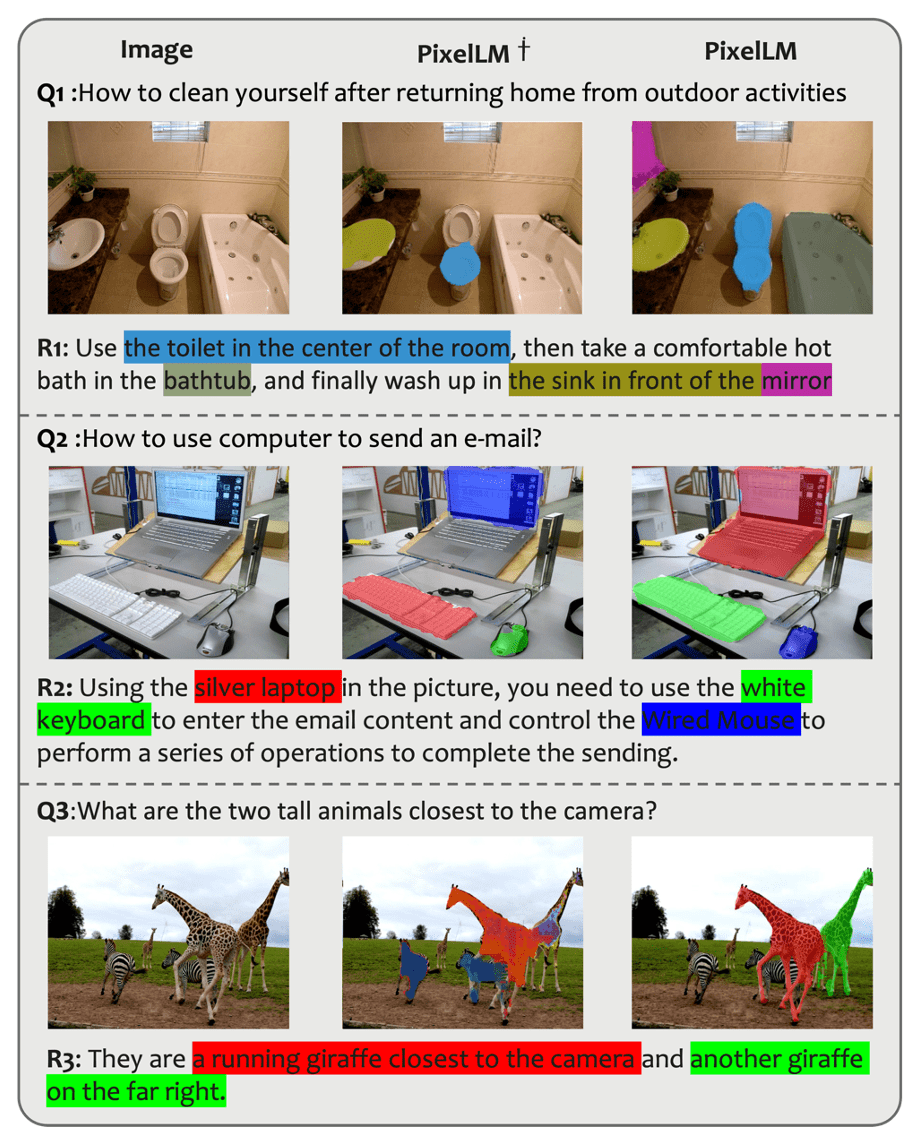}
\centering
\vspace{-1em}
\caption{\textbf{Comparison between PixelLM and PixelLM$^\dagger$} (w/o token fusion mechanism and target refinement loss.)}
\label{fig:results_comp}
\vspace{-2.2em}
\end{figure}

\vspace{-3pt}
\subsection{Ablation Study}
\label{subsec:ablation}
\vspace{-1pt}


\noindent \textbf{Number of scales.}  Tab.~\ref{tab:abla_arch} shows the effects of increasing the number scales in PixelLM by adding more feature layers and corresponding codebook tokens to the decoder. We observe notable gains by adding just one extra scale (the 2nd row). The gain diminishes with more scales. Using a larger laion-CLIP-H can further improves performance.

\noindent \textbf{Token fusion mechanism} 
In this ablation, the number of tokens in a group is 3. Table~\ref{tab:abla_method} shows that applying token fusion achieves up to a 2.4\% and 2.2\% increase in cIoU on MUSE val and refCOCOg, respectively. This demonstrates its benefits for both multi-target and single-target tasks.

\noindent \textbf{Token number in each scale} 
Tab.~\ref{tab:abla_segnum} explores how changing the number of tokens per group (denoted as 
$N$  in Section~\ref{subsec:arch}) affects performance.  $N=1$ corresponds to no token fusion.  Increasing $N$ to two and three consistently boosts performance, yielding increases of 0.9\% to 2.1\% in cIoU and 1.3\% to 2.1\% in gIoU. However, further increases yield only slight improvements in metrics.

\noindent \textbf{Target refinement loss}
As Table~\ref{tab:abla_method} shows, target refinement loss results in a 0.8\% improvement in gIoU and 0.6\% in cIoU on MUSE val. This loss, designed for scenarios with multiple targets, does not affect performance in refCOCOg. Combining the loss with token fusion can lead to up to a 2.7\% improvement in cIoU on MUSE. Fig.~\ref{fig:results_comp} compares the use (third column) and non-use (second column) of these two designs, wherein the former shows higher mask quality and better target completeness than the former.

\noindent \textbf{MUSE enhances pixel reasoning.} To further verify this, We test LISA both with and without MUSE training on ReasonSeg, a single-object reasoning dataset proposed by LISA. Training with MUSE markedly improves LISA's performance on ReasonSeg val (gIoU: 52.9 $\rightarrow$58.0, cIoU: 54.0$\rightarrow$59.1), underscoring MUSE's effectiveness in enhancing pixel reasoning. 

\noindent \textbf{GPT-4V \vs GPT-4} 
To showcase GPT-4V's advantage over GPT-4 in data generation, we create 30,000 extra multi-target question-answer pairs for a balanced comparison.
As one can see from Tab.~\ref{tab:abla_gpt}, GPT-4V-generated data shows superior performance in the MUSE benchmark, with up to 7.0\% gIoU and 8.6\% cIoU gains using the same data amount. Additionally, increasing MUSE training data volume leads to consistent performance improvements, indicating that enhancing the MUSE dataset allows PixelLM to gradually improve its general pixel-level understanding.

\vspace{-5pt}
\section{Conclusion}
\vspace{-5pt}
\label{sec:conclusion}
In this study, we introduce PixelLM, an effective and efficient LMM for pixel-level image reasoning and understanding. Benefiting from novel designs,  PixelLM is adept at producing high-quality masks for a variety of tasks. Moreover, we construct a comprehensive multi-target reasoning segmentation benchmark to bolster this research area. Through extensive experiments, PixelLM achieves promising results across various benchmarks. Future endeavors will focus on the expansion and enhancement of PixelLM's capabilities.

\section*{Acknowledgments}
This research was funded by the Fundamental Research Funds for the Central Universities (2023JBZD003), and the National NSF of China (No.U23A20314).

{
    \small
    \bibliographystyle{ieeenat_fullname}
    \bibliography{main}
}

\clearpage
\newpage
\appendix
\maketitlesupplementary

\renewcommand\thefigure{\thesection.\arabic{figure}}
\renewcommand\thetable{\thesection.\arabic{table}}
\setcounter{figure}{0} 
\setcounter{table}{0} 
\setcounter{table}{0}
\renewcommand{\thetable}{A.\arabic{table}}

In this supplementary material, we first detail the training configuration, decoder structure, flops calculation, and MUSE evaluation process in Sec.\ref{sec:supp_implement}. We then present an additional experimental analysis of our segmentation codebook and decoder in Sec.\ref{sec:supp_exp}. Furthermore, we offer a more comprehensive analysis of MUSE and the multi-referring segmentation dataset in Sec.~\ref{sec:sup_muse}.

\section{Implementation Details}
\label{sec:supp_implement}
\noindent \textbf{Training details.} 
We give the detailed training configuration in Tab.~\ref{table:supp_training_setting}, and we do \textit{not} use color jittering, drop path or gradient clip. The gradient accumulation step is set to 10.

\noindent \textbf{Structure of pixel decoder.} 
The decoder can be divided into three parts based on their functions: \textit{i}) the attention block for each scale; \textit{ii}) using the output mask from one scale to modulate the features in the next scale; \textit{iii}) the fusion of masks from all scales to obtain the final results. We present the PyTorch-style pseudocodes for the overall decoder and each part in Alg.~\ref{alg:code}.



\noindent \textbf{Calculation of TFLOPs.} 
We compare models' TFlops (trillion floating-point operations per second) in Tab. 1.  The calculation follows the formula in \cite{clark2020electra} and the script inDeepSpeed~\cite{deepspeed}. Since the flops for LMMs vary with the generated token length, we standardize it at 512. This length aligns with the common default used by~\cite{liu2023llava,touvron2023llama} and suffices to accommodate dozens of target objects. 

\noindent \textbf{Details of the evaluation metric.}
Sec.~4.3 provides a concise overview of the MUSE 
evaluation pipeline. In this section, we delve into a more formal and detailed explanation of its design. Let us denote by $M=\{M_g\}_{g=1}^G$ the ground truth set of $G$ objects, and $\hat{M}=\{\hat{M_k}\}_{k=1}^K$ the set of $K$ predictions. Motivated by ~\cite{detr}, assuming $K$ is not equal to $G$, we use $\varnothing$ (no objects) to pad the smaller set and both sets finally have size $P={\rm max}(G, K)$. 

\noindent\textbf{(1)} We find a bipartite matching between these two sets by searching for a permutation of $P$ elements, $\sigma\in\mathfrak{S}_P$, with the lowest cost:
\begin{equation*}
    \hat{\sigma} = \mathop{\arg\min}\limits_{\sigma\in\mathfrak{S}_P} \sum_{i}^{P} \mathcal{L}_{match}(M_i, \hat{M}_{\sigma(i)})  
\end{equation*}
where $\mathcal{L}_{match}(M_i, \hat{M}{\sigma(i)})$ is a pairwise matching cost between ground truth $M_i$ and a prediction with index $\sigma(i)$. We compute this optimal assignment efficiently with the Hungarian algorithm. We define $\mathcal{L}_{match}(M_i, \hat{M}{\sigma(i)})$ as $\mathcal{L}_{bce}(M_i, \hat{M}{\sigma(i)}) + \mathcal{L}_{dice}(M_i, \hat{M}{\sigma(i)})$.

\noindent\textbf{(2)} Based on the matching results, we modify the generated response $y_{res}$ to $y_{res}^{\prime}$: since each $\hat{M}_i$ originates from a segmentation token sequence in $y_{res}$, we replace each sequence with the GPT-generated description of $M_i$. 

\begin{figure*}[t]
\includegraphics[width=1\linewidth]{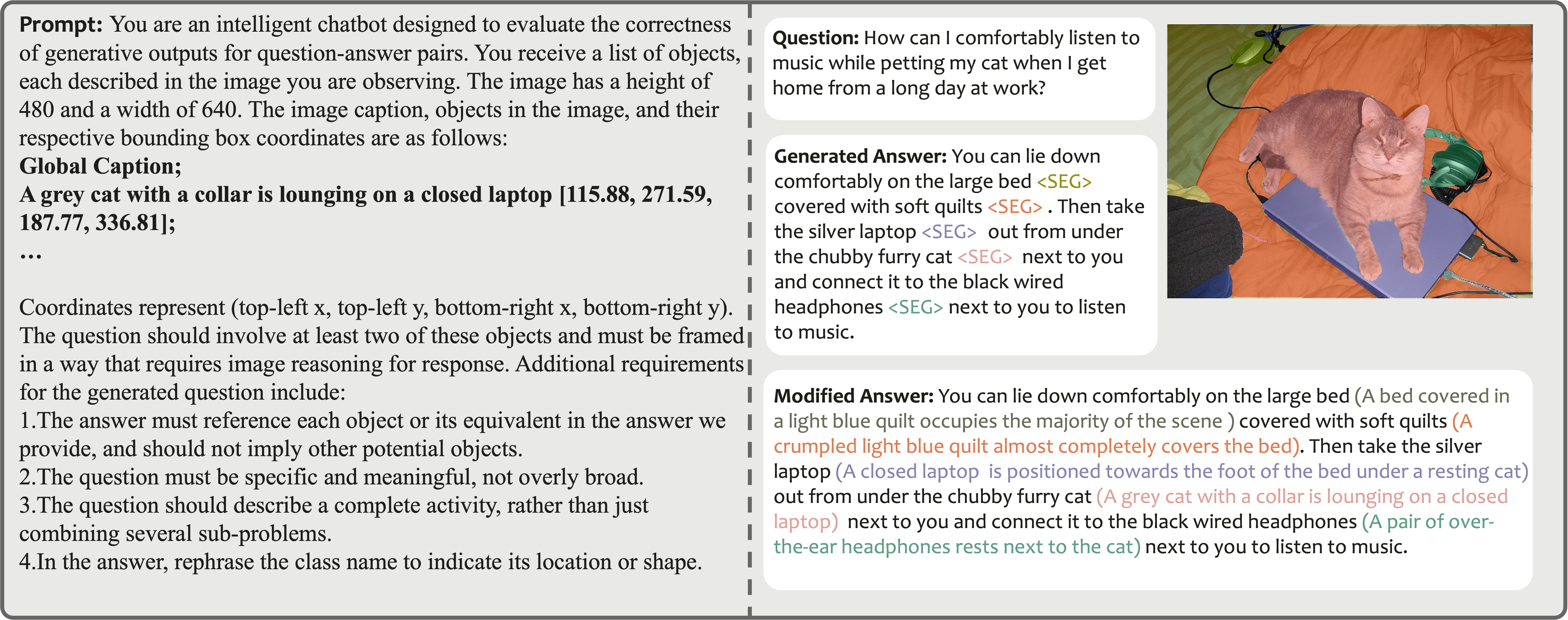}
\centering
\vspace{-3pt}
\caption{\textbf{Evaluation example.} The left panel illustrates the prompt employed in our GPT evaluation pipeline. The right panel showcases an example of a predicted answer alongside its corresponding modified version, as input to GPT.}
\label{fig:supp_evaluation}
\end{figure*}

\noindent\textbf{(3)} We use a carefully designed prompt for GPT-3.5 to assign a score $s_i$ to each $\hat{M}_i$ in the answer in a single step. An example of this methodology is depicted in Fig.~\ref{fig:supp_evaluation}. The empty predictions are directly scored with 0. 

The above three steps assess the model's capability to generate outputs where masks are intertwined with text descriptions and evaluate how accurately these masks correspond to their respective text descriptions. Then we evaluate the quality of the masks.

\begin{table}[t]
    \footnotesize
    \centering
    {
\begin{tabular}{p{3.4cm}|p{2.2cm}}
     config & value\\

    \shline
    optimizer & AdamW\\
     base learning rate & 3.0e-4\\
     weight decay & 0\\
     optimizer monmentum & $\beta_1$, $\beta_2$=0.9,0.95 \\
     batch size & 16 \\
     learning rate schedule & WarmipDecayLR \\
     warmup iterations & 100 \\
     augmentations & None \\
     $\alpha$ & 2.0 \\
     $\lambda_{ref}$ & 2.0 \\
     $\lambda_{dice}$ & 0.5 \\

\end{tabular}
}
\caption{\textbf{Training settings.}}
\label{table:supp_training_setting}
\vspace{-1em}
\end{table}
\begin{table}[t]
    \footnotesize
    \centering
    \resizebox{\columnwidth}{!}
    {
\begin{tabular}{l | cc | ccc|cc}
    \toprule
    \multirow{2}*{\makecell{Codebook \\design}} &\multicolumn{2}{c|}{MUSE Val} &\multicolumn{3}{c|}{refCOCO+} &\multicolumn{2}{c}{refCOCOg}  \\
     & gIoU & cIoU & val & testA &testB  &val(U) & test(U) \\
    \midrule

    $N$ &41.0  &48.3  &64.0 &69.8 &57.5  &67.9 &68.4\\

    $N\times L$ &\textbf{42.6}  &\textbf{50.7} &\textbf{66.3}&\textbf{71.7}&\textbf{58.3} &\textbf{69.3} &\textbf{70.5}\\

    \bottomrule
    
\end{tabular}
}
\caption{\textbf{Sharing tokens across $L$ scales.} The first row corresponds to results of sharing tokens across all feature scales.}
\label{table:supp_scale_share}

\end{table}
\begin{table}[t]
    \footnotesize
    \centering
    \resizebox{\columnwidth}{!}
    {
\begin{tabular}{l | cc | ccc|cc}
    \toprule
    \multirow{2}*{\makecell{Selected layers}} &\multicolumn{2}{c|}{MUSE Val} &\multicolumn{3}{c|}{refCOCO+} &\multicolumn{2}{c}{refCOCOg}  \\
     & gIoU & cIoU & val & testA &testB  &val(U) & test(U) \\
    \midrule
    baseline &40.1  &47.2  &61.1&65.4&54.7 &64.3 &65.6\\
    20, 17, 14 &42.0  &51.5 &66.0&71.4 &58.5  &68.8 &70.6\\
    20, 17 &41.2  &49.1 &65.9&71.2&58.0 &68.0 &66.4\\
    
    23, 14, 10  &41.8  &48.0 &65.1 &68.3&57.9 &67.5 &68.0\\
    23, 14, 20 &42.3  &\textbf{51.4}  &66.0 &71.5 &\textbf{58.5}  &\textbf{69.8} &70.4\\
    23, 14 &\textbf{42.6}  &50.7  &\textbf{66.3}& \textbf{71.7} &58.3 &69.3 &\textbf{70.5}\\

    \bottomrule
    
\end{tabular}
}
\caption{\textbf{Multi-scale layer selection.} CLIP-ViT consists of 24 layers. ``Baseline" only uses the penultimate (i.e., the 23rd) layer.}
\label{table:supp_scale_selection}
\vspace{-2em}
\end{table}

\noindent\textbf{(4)} The final IoU of each prediction is:
\begin{equation*}
\begin{aligned}
{\rm Intersection}_i &= \left\{ 
    \begin{array}{lc}
        {\rm Intersection}_i &s_i>0.5 \\
        0 &s_i\leq0.5
    \end{array} \right. \\
{\rm IoU}_i &= {\rm Intersection}_i / {\rm Union}_i
\end{aligned} 
\end{equation*}
And the final IoU$_{img}$ of each image is:
\begin{equation*}
{\rm IoU}_{img} = \sum\nolimits_i {\rm IoU}_i / P
\end{equation*}
Based on the IoU scores, we can calculate gIoU and cIoU metrics by referring segmentation dataset.

    \begin{algorithm}[t]

      \caption{Pseudo codes of our pixel decoder.}
      \label{alg:code}
      \scriptsize
          \Comment{Inputs:~f\_img:image features from L scales [L,C,H,W]; 
          h\_seg:segmentation tokens for L scales [L, C];}

          \Comment{Variables:~lev\_token:~Learnable embeddings for L scales [L,C]; out\_token:Learnable embeddings [N, C]; image\_pe:position embedding of image features; 
          gamma:mask weighting factors [L]}
          \Comment{Functions:~SelfAttention();CrossAttention();MLP();
          up\_scale(); down\_scale();}

    \Function{feature\_update(f, mask)}{
    \Comment{f:image feature of one scale.[CxHxW]}
    \Comment{mask:output from the attention block above f.[Cx4Hx4W]}

     \var{mask = down\_scale(mask, size=(HxW))}; \Comment{mask: [H,W]}
     
    \var{f = f * (sigmoid(mask) + 1)} \Comment{update feature}

    \var{return f}
    
  }
  \Function{attention\_block(h, f, l)}{
     \Comment{h, f:segmentation token and image feature of one scale.[C], [CxHxW]}
     \Comment{l:scale index}

     \var{token = cat([out\_token, h], dim=0) + lev\_token[l]}; \Comment{token:[N+1,C]}

     \var{attn\_out = SelfAttention(q=k=v=token)};\Comment{self attention output:[N+1,C] }
     
     \var{token = norm(token + attn\_out)}

     \var{key = f + image\_pe}; \Comment{key:[C,H,W]}

     \var{attn\_out = CrossAttention(q=token, k=key, v=f)};\Comment{cross attention output:[N+1,C] }

     \var{token = norm(token + attn\_out + MLP(token)) };\Comment{update token:[N+1,C] }

    \var{attn\_out = CrossAttention(q=key, k=token, v=token)};\Comment{cross attention output:[C,H,W]} 
    
    \var{f = norm(f + attn\_out)};  \Comment{update feature:[C,H,W]}

    \var{f\_up = up\_scale(f)} \Comment{f\_up:Cx4Hx4W}
    \var{token = MLP(token)} \Comment{token:[N+1,C]}
    \var{mask = token @ f} \Comment{mask:[N+1,4H,4W]}
    \var{mask = mean(mask, dim=0)} \Comment{mask:[4H,4W]}
    
     \var{return mask}
     }
  \Function{mask\_fusion(mask\_list)}{
     \Comment{masks:a list of masks from each scale}

    \var{final\_mask = zeros(4Hx4W)}; \Comment{initialize empty mask}
    
    \For{l, m in enumerate(mask\_list)}{
    \var{final\_mask = final\_mask + gamma[l] * m}
    }

    \var{return final\_mask}
  }

  \Comment{Main Function}
  \Function{pixel\_decoder(f\_img, h\_seg)}{
     \Comment{masks:a list of mask from each scale}

    \var{mask\_list = []}
    
    \For{l, (f, h) in enumerate(zip(f\_img, h\_seg))}{
    
    \If{l < L-1}{
    
    \var{f = feature\_update(f, mask)}
    }; \Comment{update features after the scale L}
    
    \var{ mask = attention\_block(h, f, l)}
    
    \var{mask\_list.append(mask)}
    }

    \var{final\_mask = mask\_fusion(mask\_list)}
    
    \var{return final\_mask}
  }

    \end{algorithm}

\begin{figure*}[t]\centering
  \subfloat[\label{fig:cat_number}The number of instances per category]{\includegraphics[width=0.32\linewidth]{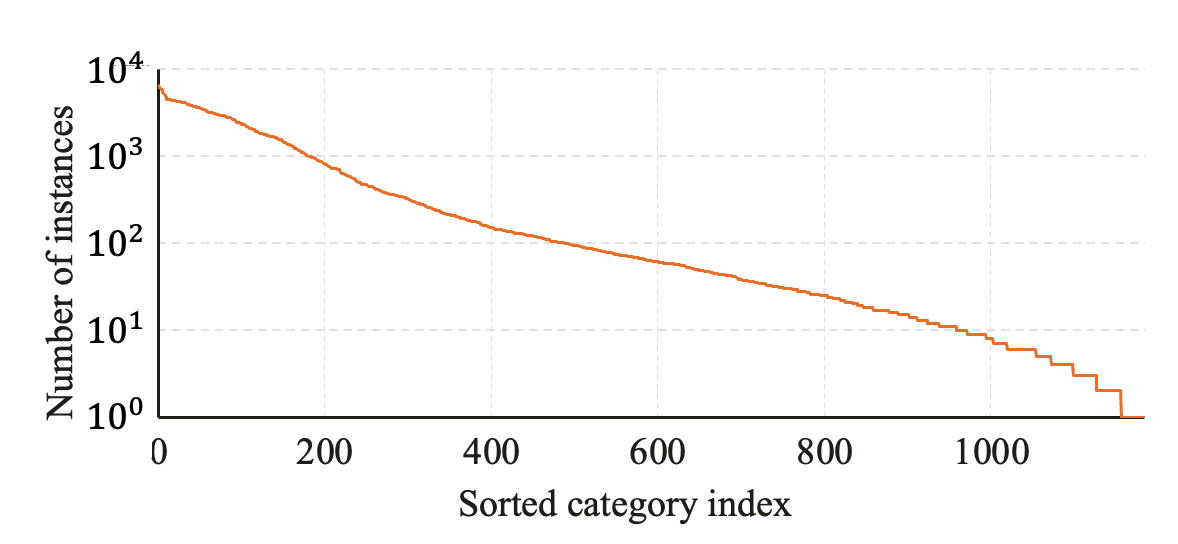}  }
\hfill
\subfloat[\label{fig:token_number}Distribution of token count in instance descriptions.]{\includegraphics[width=0.32\linewidth]{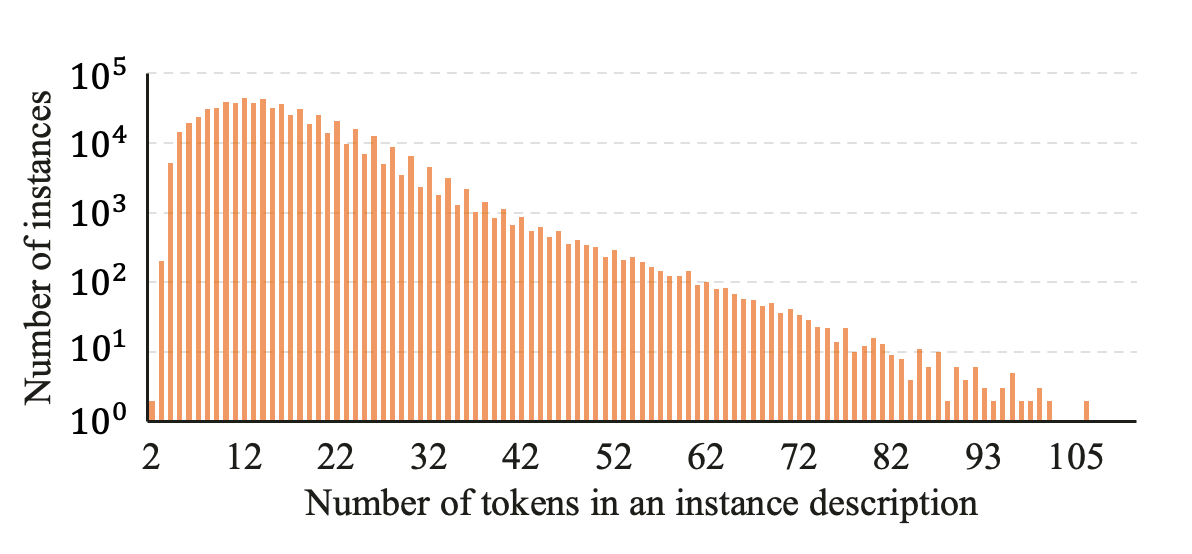} }
\hfill
\subfloat[\label{fig:target_number} Distribution of target count in questions.]{\includegraphics[width=0.32\linewidth]{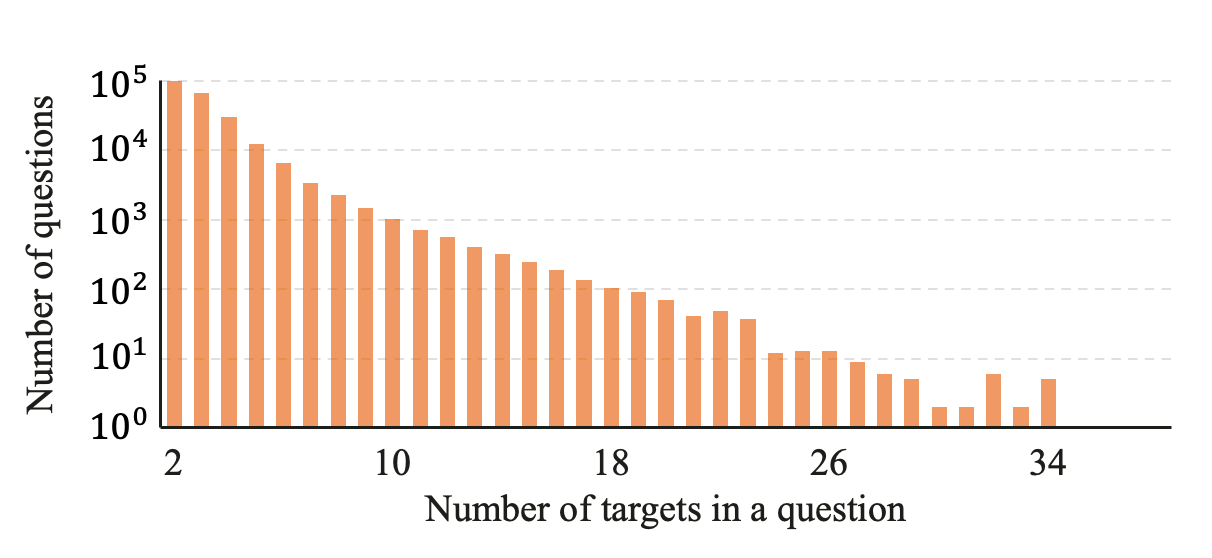}}\\[.25em]
\caption{\textbf{Dataset statistics}. Best viewed digitally.}
\vspace{-3mm}
\label{fig:supp_data_comp}
\end{figure*}

\section{More Ablative Experiments.}
\label{sec:supp_exp}
 
\noindent \textbf{Multi-scale tokens sharing.} To clearly demonstrate the necessity of our multi-scale segmentation tokens in the codebook $C_{\text{seg}}$, we continue using $L$ multi-scale features, but reduce the original codebook shape from $N\times L$ to $N$ (recall that $N$ is the number of tokens in each scale group). This reduction resulted in the remaining $N$ tokens being identical (i.e., shared) across all $L$ scales. As Tab.~\ref{table:supp_scale_share} shows, using a dedicated segmentation token for each scale yields better performance.

\noindent \textbf{Multi-scale layer selection.} We experiment with different layer selections as presented in Tab.~\ref{table:supp_scale_selection}. The CLIP-ViT-L used in PixelLM consists of 24 layers. To reduce the computational cost of the decoder, we avoid selecting features from all layers of ViT. Instead, we follow the multi-scale feature selection ratio from prior work~\cite{he2015deep,liu2021swin} and consider other selection options. Our experiments reveal that selecting features from layers before the middle does not yield benefits for our task (the fourth row in Tab.~\ref{table:supp_scale_selection}). Therefore, we mainly focus on selections from the middle and rear layers. The results show that using features from layers 14 and 23 leads to the best outcomes.


\section{More Details about MUSE}
\label{sec:sup_muse}

\subsection{Data Filtering}
\label{subsec:supp_data_filter}
\noindent \textbf{GPT-4V filtering.} 
Although GPT-4V can efficiently understand image content, there are still failure cases in the generated data, which can be summarized in the following two points: 
\begin{itemize}
    \item Questions are vague and open to multiple interpretations. For example, the question ``What should I take with me on my outing?'' is extremely vague because ``outing'' can mean a wide array of activities.
    \item Answers that omit semantically equivalent instances. For instance, a question might ask about ``choosing a fruit for a snack'', but the answer may only suggest ``an apple'', ignoring other fruit visible in the image.
\end{itemize}
Therefore, it is necessary to employ a stringent sample filtering process to guarantee the quality of the data. Toward this goal, we develop a GPT-4V assisted data filtering pipeline. This pipeline operates by prompting GPT-4V to evaluate all initial question-answer pairs, based on identified common failure modes. Pairs classified by GPT-4V as falling within these failure categories are removed. This procedure effectively excludes approximately 20\% of the preliminary data. The specific prompts used in this process are detailed in Fig.~\ref{fig:supp_evaluation}.

\noindent \textbf{Human verification.} To ensure high quality in the generated question-answer pairs during the evaluation stage, we further engage experienced human annotators to double-check our evaluation set. Our approach is driven by two primary objectives:
\begin{itemize}
    \item The questions should follow an intuitive and logical sequence that a person would typically think of when viewing the image.
    \item The answers should correspond closely to the way a human would naturally respond to the question
\end{itemize}
The above filtering process effectively ensures that the questions in the MUSE dataset are sufficiently challenging for reasoning, while the answers remain detailed and accurate.

\begin{figure*}[t]
\includegraphics[width=1\linewidth]{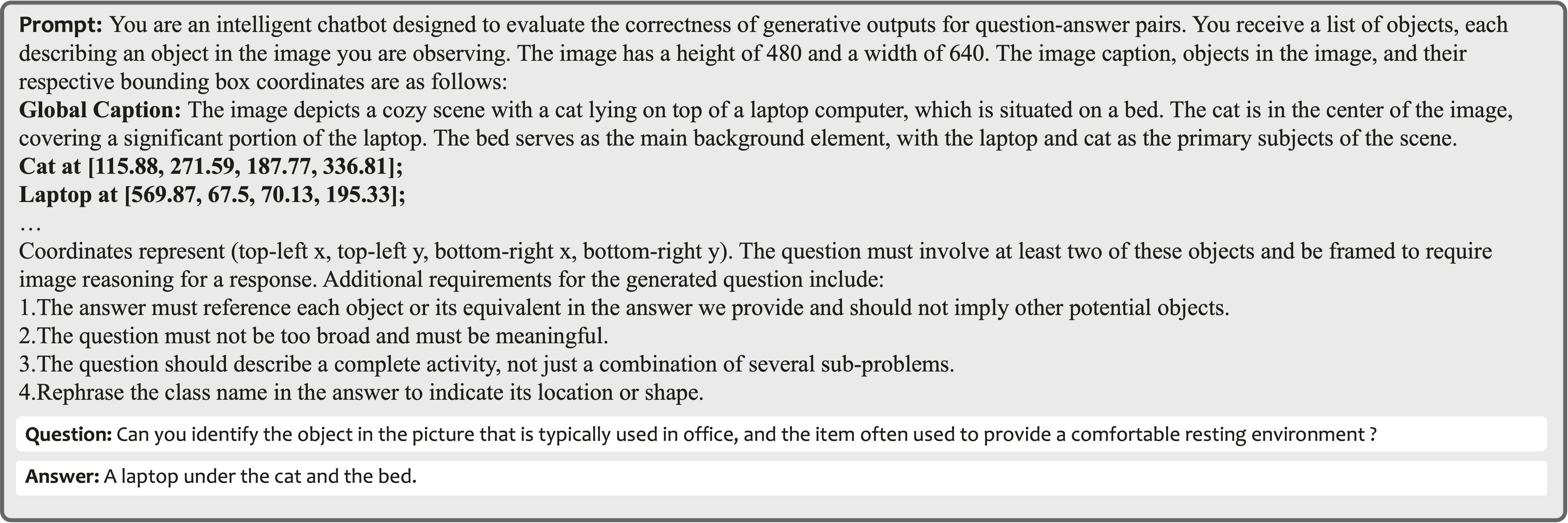}
\centering
\vspace{-3pt}
\caption{\textbf{Example of GPT-4 data generation.} The corresponding image is the same as in Fig.~\ref{fig:supp_evaluation}.}
\label{fig:supp_gpt4_gen}
\end{figure*}

\subsection{Dataset Statistics}
\label{subsec:data_ana}
In this section, we systematically analyze our dataset. First, our question-answer pairs are based on over 1000 categories, encompassing a wide spectrum of objects found in daily scenes. Additionally, the descriptions of objects in the answers go beyond mere category names limited to a few tokens. Instead, they offer context-specific descriptions extending to over 100 tokens. This demonstrates that our dataset is rich in perception information, crucial for real-world applications.
Finally, we present the statistics regarding the number of objects involved in a data sample. 

\noindent \textbf{Category statistics.} There are over 1000 categories in MUSE from the original LVIS dataset, and 0.9 million instances with unique descriptions that vary based on the context of the question-answer pairs. Fig.~\ref{fig:cat_number} shows the number of instances per category on all question-answer pairs. The distribution inherits the low-shot nature of LVIS.  

\noindent \textbf{Token count.} Fig.~\ref{fig:token_number} presents the distribution of instances by token count in their descriptions, highlighting a wide range that exceeds 100 tokens in the most extensive cases. These descriptions are not limited to simple category names; rather, they are substantially enriched with detailed information about each instance, encompassing aspects like appearance, attributes, and relationships with other objects, thanks to our GPT-4V-based data generation pipeline. The depth and variety of information in the dataset bolster the trained model's generalization capabilities, enabling it to effectively address open-set questions.

\noindent \textbf{Target count.} Fig.~\ref{fig:target_number} presents statistics on the number of targets in each question-answer pair. The average number of targets is 3.7, with the maximum number of targets in a single pair reaching up to 34. This number can cover most scenarios of target reasoning for a single image.


\subsection{Multi-referring Segmentation}
\label{subsec:multi_referring}
As mentioned in Sec.~5.1, we transform conventional referring segmentation datasets into a multi-referring format for model training. In this subsection, we detail this process. The transformation involves selecting one to three distinct target objects from the annotations of each image. These objects are used to construct questions in the format: \texttt{Please segment the $<$objects$>$ in the image}, where {\tt $<$objects$>$} represents a list of comma-separated object descriptors. The response format is a list of comma-separated {\tt $<$object$>$ is $<$SEG$>$}, with {\tt $<$object$>$} being the description of each object. We require that the order of predictions in the answer matches the order of object names in the question and calculate gIoU and cIoU for each prediction based on this.


\subsection{More Details about GPT-4 Generated Data.}
\label{subsec:supp_gpt4}
In Sec.~5.5, we create 30,000 additional multi-target question-answer pairs to compare GPT-4 and GPT-4V. The prompts for GPT-4 adhere to similar generation principles but incorporate detailed image captions to offset the absence of visual content. Fig.~\ref{fig:supp_gpt4_gen} demonstrates an example of our GPT-4 data generation process and its typical failures. Given that image content is not directly perceivable, conveying as detailed an image description as possible into GPT-4 is crucial to compensate for this information gap. However, this method often leads to a lack of diversity in the generated question-answer pairs (refer to the question-answer pair at the bottom of Fig.~\ref{fig:supp_gpt4_gen}), characterized by:  \textit{i}) Numerous questions are composed of simple referring style sub-questions; \textit{ii}) The question and answer content is limited to what the image caption describes; \textit{iii}) Challenges in generating detailed object descriptions. To address these issues, it might be necessary to introduce more complex annotation details like object relationships, which significantly increases the complexity and burden of data generation.

\end{document}